\def\year{2017}\relax
\documentclass[letterpaper]{article}
\usepackage{aaai17}
\usepackage{times}
\usepackage{boldline}

\usepackage{graphicx,wrapfig}
\usepackage{enumitem}
\usepackage{amsmath,amsbsy,amsfonts,amssymb,amsthm}
\usepackage{algorithm,algorithmic,mathtools}
\usepackage{color,cases}
\usepackage{tikz,bbm}
\usetikzlibrary{calc,shapes}
\usepackage{subfigure}
\usepackage{comment}
\usepackage{xcolor,url,verbatim}

\usepackage{helvet}
\usepackage{courier}
\DeclareMathOperator{\dist}{dist}

\newcommand{\abs}[1]{\left\lvert #1 \right\rvert}
\newcommand{\paran}[1]{\left( #1 \right)}

\newcommand{\trnorm}[1]{\left\lVert #1 \right\rVert_{*}}
\newcommand{\frobnorm}[1]{\left\lVert #1 \right\rVert_{F}}
\newcommand{\infnorm}[1]{\left\lVert #1 \right\rVert_{\infty}}
\newcommand{\nn}{\nonumber}

\newtheorem{proposition}{Proposition}
\newtheorem{lemma}{Lemma}
\newtheorem{theorem}{Theorem}
\newtheorem{remark}{Remark}
\newtheorem{corollary}{Corollary}

\newcommand{\se}{\widehat{\sigma}}
\newcommand{\pe}{\widehat{P}}
\newcommand{\proj}{\mathcal{R}}
\newcommand{\pec}{\overline{\widehat{P}}}

\newcommand{\R}{\mathbb{R}}

\renewcommand{\P}{\mathcal{P}}
\renewcommand{\S}{\mathcal{S}}

\begin{document}
\title{Inductive Pairwise Ranking: Going Beyond the $n \log(n)$ Barrier}
\author{
U.N. Niranjan~\thanks{Part of the work done while interning at Xerox Research Centre India.} \\
Department of Computer Science \\
University of California Irvine \\
un.niranjan@uci.edu \\
\And
Arun Rajkumar \\
Data Analytics Lab \\
Xerox Research Centre India \\
arun.rajkumar@xerox.com \\
}
\maketitle
\begin{abstract}
We study the problem of ranking a set of items from non-actively chosen pairwise preferences where each item has feature information with it. We propose and characterize a very broad class of preference matrices giving rise to the \textit{Feature Low Rank} (FLR) model, which subsumes several models ranging from the classic Bradley--Terry--Luce (BTL)~\cite{bradley1952rank} and Thurstone~\cite{thurstone1927law} models to the recently proposed blade-chest~\cite{chen2016modeling} and generic low-rank preference~\cite{rajkumar2016can} models. We use the technique of matrix completion in the presence of side information to develop the Inductive Pairwise Ranking (IPR) algorithm that provably learns a good ranking under the FLR model, in a sample-efficient manner. In practice, through systematic synthetic simulations, we confirm our theoretical findings regarding improvements in the sample complexity due to the use of feature information. Moreover, on popular real-world preference learning datasets, with as less as 10\% sampling of the pairwise comparisons, our method recovers a good ranking.
\end{abstract}

\section{Introduction}
Ranking from pairwise comparisons or \textit{preferences} is an ubiquitous problem in machine learning, statistics and theoretical computer science. In the so-called \textit{non-active} setting, one is given comparison results of $m$ pairs pre-selected from among all pairs of $n$ items where each pair is compared at least $K$ times. Particularly, the learner does not get to choose which pairs are to be compared. The goal is then to estimate a suitable ordering of the items, using the observed comparison results, that conforms to the true ordering, assuming one exists, up to the desired error $\epsilon$ in a suitably defined error measure.

In practical ranking applications, we often have side information associated with the items that need to be ranked -- such a scenario is referred to as the \textit{inductive} setting. An advantage in this inductive learning setting is that, in addition to ranking a given set of items, one is also able to rank new unseen items that may introduced after parameter learning. Motivated by these factors, we wish to leverage the available side information to compute an ordering more efficiently than existing techniques. This is relevant in many practical applications; for instance, in addition to using a minimal amount of customer preference data, (a) using food characteristics like nutrition, preparation method, etc could help in finding the top-rated dishes of a restaurant, (b) using car features like engine type, body type, etc could help elicit useful trends for automotive industry.
\paragraph{Our Contributions: }To the best of our knowledge, our work is the first to derive a provable and efficient method for ranking in the non-active inductive setting. Our novelty and technical contributions can be summarized along the following axes:
\begin{enumerate}
\item \textit{Model: }We generalize existing models so that we can incorporate (a) features, and (b) feature correlations associated with the items to be ranked. We show that our model subsumes many existing and popular ranking models.
\item \textit{Algorithm: }Our algorithm uses two key subroutines namely, (a) noisy inductive matrix completion, and (b) approximate pairwise ranking algorithm~\cite{copeland1951reasonable}.
\item \textit{Guarantee: }We derive the guarantee that our algorithm obtains, with high probability, an $\epsilon$-accurate recovery using $\Omega(\max(\log n / \epsilon^2, d^4 \log^3 n / \epsilon^3 n^2))$ independent pairwise comparisons chosen uniformly at random.
\item \textit{Experiments: }We substantiate our theoretical results by demonstrating sample complexity gains on both synthetic and real-world experiments.
\end{enumerate}
We would like to emphasize upfront that it is the sole focus of this paper to study the practically motivated regime of $d \ll n$ in detail. Furthermore, we note that our sample complexity results do not violate the standard $\Omega(n \log n)$ lower bounds for comparison-based sorting algorithms since we develop an algorithm that effectively ranks in the \textit{feature space} rather than the \textit{item space}.

\subsection{Related Work and Background}
We now give a brief overview of relevant work in ranking models followed by a brief background regarding tools from inductive matrix completion theory which will be crucial in proving our sample complexity bounds.
\paragraph{Ranking Models :}In the simplest terms, the ranking problem involves estimating the \textit{best} ordering items according to some observed preferences. A early thread of ranking literature has its beginnings in economics involving choice models~\cite{r1959individual}; other related works in social choice theory include \cite{lu2011learning} and \cite{caragiannis2013noisy}. A certain deterministic version of the ranking problem is also studied as the \textit{sorting} problem which is central in theoretical computer science.
\begin{enumerate}
\item \textit{Random Utility (RU) models: }Starting with the seminal work of \cite{bradley1952rank}, the Bradley-Terry-Luce (BTL) model has become a landmark model for ranking. In the vanilla version of this model, the probability that item $i$ beats item $j$ is given by $P_{ij} = \frac{w_i}{w_i + w_j}$ where $\mathbf{w} \in \R^n_+$ is the parameter vector to be estimated from data; the $i^{th}$ entry in $w$ denotes the \textit{score} associated with item $i$. Thurstone~\cite{thurstone1927law} model is also a well-known statistical model; here, $P_{ij} = \Phi (s_i-s_j)$ where $\Phi$ is the standard normal Cumulative Distribution Function (CDF) and $\mathbf{s} \in \R^n$ is the score vector. These classic models fall under the so-called \textit{Random Utility} (RU) Models~\cite{marschak1959binary}.
\item \textit{Item Feature (IF) models: }Extending the BTL model, statistical models that utilize side information are presented in \cite{cattelan2012models}. Recently, \cite{chen2016modeling} presented the blade-chest ranking model which studied the stochastic intransitive setting. Their algorithm involves regularized maximum likelihood estimation for which tight sample complexity properties are not known. Despite the above works, to the best of our knowledge, there are no known models utilizing feature information while have provable sample-efficient algorithms for estimation and ranking.
\item \textit{Low Rank (LR) models: }Recently, \cite{rajkumar2016can} -- unifying classic models such as BTL and Thurstone models -- defined a generic class of preference matrices which have low rank under transformations involving suitable \textit{link functions}. Upon such a transformation, connections of the ranking problem to matrix completion theory become clear. Subsequently, they use well-known matrix completion results to derive sample complexity guarantees for ranking. However, their model does not utilize side information that may be available.
\end{enumerate}
This list is by no means exhaustive; while there exist several other ranking methods (eg, ranking-SVM\cite{joachims2002optimizing}), there are no known sample complexity guarantees associated with these.

\paragraph{Inductive Matrix Completion: }The matrix completion task~\cite{candes2009exact} is to fill-in the missing entries of a partially observed matrix, which is possible efficiently under a low-rank assumption on the underlying matrix. Oftentimes, side information may be available which further makes this task potentially easier. This is the \textit{Inductive Matrix Completion} (IMC) problem which is formally defined as the optimization problem, 
$\widehat{\mathbf{Z}} = \arg \min_{\mathbf{Z}} \ell( (A^\top Z B)_{ij}, M_{ij})$ 
where $\mathbf{A} \in \R^{d_1 \times n_1}$ and $\mathbf{B} \in \R^{d_2 \times n_2}$ are known feature matrices, $\mathbf{Z} \in \R^{d_1 \times d_2}$ is a rank-$r$ unknown latent parameter matrix, $(i,j) \in \Xi \subseteq [n] \times [n]$ is the support set corresponding to the (uniformly sampled) observed entries and $\ell$ is any loss function, the squared loss being the most commonly chosen one. Once the estimate $\widehat{\mathbf{Z}}$ is obtained using the training set indexed by $\Xi$, predictions may then be performed as $\widehat{M}_{ij} = (A^\top \widehat{Z} B)_{ij}$ for any $(i,j) \in \Xi^c$. The known solution techniques with recovery guarantees are:
\begin{enumerate}
\item \textit{Non-convex algorithm (via alternating minimization): }This approach entails parameterizing $\mathbf{Z} = \mathbf{U} \mathbf{V}^\top$ and performing alternating projected least squares updates on $\mathbf{U}$ and $\mathbf{V}$. The tightest known guarantee for this approach involves a sample complexity of $\Omega(d^2 r^3 \kappa^2 \log(d))$ and a convergence rate of $O(\log(1/\epsilon))$~\cite{jain2013provable}.
\item \textit{Convex relaxation (via trace-norm formulation): }This approach entails relaxing the rank constraints to a nuclear norm penalty. Existence of a unique optimum can be shown with high probability \cite{xu2013speedup} and is characterized a sample complexity of $\Omega(d r \log(d) \log(n))$. Despite the non-smoothness, a sub-gradient descent algorithm provably converges with a rate of $O(1/\sqrt{\epsilon})$~\cite{ji2009accelerated}. Noisy features are handled in \cite{chiang2015matrix}.
\end{enumerate}

\section{Feature-aware Ranking}
\subsection{Notation and Preliminaries}
\textit{General notation: }Unless stated otherwise, we use lower-case letters for scalars, upper-case letters for universal constants, lower-case bold-face letters for vectors and upper-case bold-face letters for matrices; specifically, $\mathbf{P}$ denotes a preference matrix. For any matrix $\mathbf{M} \in \R^{a \times b}$, let $\infnorm{\mathbf{M}} = \max_{i,j} \abs{M_{ij}}$, $\trnorm{\mathbf{M}} = \sum_{i=1}^{\min \{ a,b \}} \sigma_i(\mathbf{M})$ where $\sigma_i(\mathbf{M})$ are the singular values of $\mathbf{M}$ and $\frobnorm{\mathbf{M}} = \sqrt{\sum_{i=1}^{a} \sum_{j=1}^{b} M_{ij}^2}$. $\mathbf{I}$ denotes the identity matrix whose dimensions would be implied from the context; similarly, depending on the context, $\mathbf{0}$ denotes a vector or matrix of zeros of the appropriate dimension. Next, let $P_{\min} = \min_{i \neq j} P_{ij}$ and $\Delta = \min_{i \neq j} \abs{\psi(P_{ij}) - \psi(1/2)}$. Let $\Xi$ be the support set of the observed entries of a matrix and let $m = \abs{\Xi}$. Define projection of a matrix on the support set $B = \proj_\Xi(A)$ as: $B_{ij} = A_{ij}$ if $(i,j) \in \Xi$ and $B_{ij} = 0$ if $(i,j) \notin \Xi$.  \\
\textit{Items and features: }Let $n$ be the number of items to be ranked. Let $\S_n$ denote the symmetric group on $n$ items. Let each item have a $d$-dimensional feature vector associated with it, ie, $\mathbf{f}_i \in \R^d, \forall i \in [n]$; concatenating these, we obtain the feature matrix $\mathbf{F} = [\mathbf{f}_1, \dots, \mathbf{f}_n] \in \R^{d \times n}$. \\
\textit{Link functions: }Any $\psi: [0,1] \rightarrow \R$ which is a strictly increasing bijective function is a valid link function. For example, $\psi$ could be the logit function, which is the inverse of the sigmoid function, defined as, $\psi(x) := \log \paran{\frac{x}{1-x}}$
for $x \in [0,1]$; another example is the probit function defined as $\psi(x) = \Phi^{-1} (x)$ where $\Phi$ is the standard normal CDF. When we apply the link function to a matrix, we mean that the transformation applied entry-wise. \\
\textit{Preference matrices: }Let $\P_n := \{ \mathbf{P} \in [0,1]^{n \times n} | P_{ij}+P_{ji} = 1 \}$ denote the set of all pairwise preference matrices over $n$ items. Let the set of stochastic-transitive matrices be $\P_n^{ST} := \{ \mathbf{P} \in \P_n | P_{ij} > 1/2, P_{jk} > 1/2 \implies P_{ik} > 1/2 \}$ and the set of stochastic-intransitive matrices be $\P_n^{SI} := \{ \mathbf{P} \in \P_n | P_{ij} > 1/2, P_{jk} > 1/2 \implies P_{ik} < 1/2 \}$. Let $\P_n^{RU}$ be the set of preference matrices associated with unary random utility models (which are described in the next section). Let $\P_n^{IF} := \{ \mathbf{P} \in \P_n | P_{ij} = \psi^{-1}(\mathbf{w}^\top (\mathbf{f}_i - \mathbf{f}_j)) \}$ for some $\mathbf{w} \in \R^d$.

Let $r \leq n$. Define the set of preference matrices having rank-$r$ under the link function $\psi$ as $\P_n \paran{\psi, r} := \{ \mathbf{P} \in \mathcal{P}_n | \text{ rank } (\psi(\mathbf{P})) \leq r \}$. Next, define the set of preference matrices having rank-$r$ under the link function $\psi$ with the associated feature matrix $\mathbf{A} \in \R^{d \times n}$ as $\P_n \paran{\psi, r, \mathbf{A}} := \{ \mathbf{P} \in \mathcal{P}_n \paran{\psi, r} | \psi(\mathbf{P}) = \mathbf{A}^\top \mathbf{L} \mathbf{A} \}$ where $\mathbf{L} \in \R^{d \times d}$ is an unknown rank-$r$ latent matrix (which is a function of the parameters of the ranking model) and $\mathbf{A} = [\mathbf{a}_1, \ldots, \mathbf{a}_n] \in \R^{d \times n}$ is the known feature matrix whose $i^{th}$ column is the feature vector corresponding to the $i^{th}$ item. Let $\kappa = \sigma_{\min}(\mathbf{A}) / \sigma_{\max}(\mathbf{A})$ be the inverse condition number of the feature matrix $A$. Let $i \succ_{\mathbf{P}} j$ iff $P_{ij} > 1/2$. Denoting the indicator function by $\mathbbm{1}$, we define the distance between a permutation $\sigma \in \S_n$ and a preference matrix $\mathbf{P} \in \P_n$ as:
\begin{align*}
\dist \paran{\sigma, \mathbf{P}} & := \begin{pmatrix}n \\ 2 \end{pmatrix}^{-1} \sum_{i<j} \mathbbm{1} \paran{(i \succ_{\mathbf{P}} j) \wedge (\sigma(i) \succ \sigma(j))} \\
& + \begin{pmatrix}n \\ 2 \end{pmatrix}^{-1} \sum_{i<j} \mathbbm{1} \paran{(j \succ_{\mathbf{P}} i) \wedge (\sigma(j) \succ \sigma(i))} 
\end{align*}
Note that the above distance measure essentially counts the fraction of pairs on which $\sigma$ and $\mathbf{P}$ disagree, and can be thought of as a normalized $0-1$ loss function.

\subsection{Feature Low Rank Model}
\textit{Random Utility} (RU) models, arising in discrete choice theory, dating back to \cite{marschak1959binary}, characterize the probability of an item $i$ beating item $j$, $P_{ij}$, using a prior on the (latent) score associated with those items, $w_i \in \R$ and $w_j \in \R$. The most popular pairwise ranking models including BTL and Thurstone models fit in this framework. In particular, it is well-known that if $w_i \sim Gumbel(0,1)$, we obtain the BTL model; for completeness, we justify it below:
\begin{align*}
P_{ij} = \Pr(w_i > w_j) = \Pr(w_i-w_j > 0) \stackrel{\xi_1}{=} \frac{e^{-(w_i-w_j)}}{1+e^{-(w_i-w_j)}}
\end{align*}
where $\xi_1$ follows from the fact that the difference of two independent standard Gumbel distributed random variables follows the standard logistic distribution. Similarly, if $w_i \sim \mathcal{N}(0,1)$, we obtain the Thurstone model. 
The underlying commonality in these models is the simple observation that the prior distribution is on the scores, which are unary terms. Notably, the recent result by \cite{rajkumar2016can} shows that under the inverse transformation of the CDF of the difference of the latent score variables, the preference probability matrix is low-rank for BTL and Thurstone models. Further, they extended this result to a broader class of \textit{low-rank} models in which the preference matrices are low-rank when the link function is set to be this inverse CDF.

One angle of motivation for this paper stems from the intuitive thought that the scores associated with an item $i$ in RU models can be generalized to functions involving, not just unary terms but also, pairwise terms, ie, the score of \textit{item $i$ with respect to item $j$} is given by an energy function $E_{ij}$ that has a bilinear form. From this point onwards, for simplicity, we detail the generalization of the RU models encompassing the BTL model, ie, we posit that $E_{ij}$ has a standard Gumbel distribution and consequently, we choose the link function $\psi$ to be the logit function. It is noteworthy that our results will hold under any link function for the corresponding prior.

We now propose the enery-based generative model, which we call \textit{Feature Low Rank} (FLR) model, defined via the preference matrix specified as follows:
\begin{equation}
\label{eqn:gbtl_mat}
P_{ij} = \frac{e^{-E_{ij}}}{e^{-E_{ij}}+e^{-E_{ji}}}
\end{equation}
Here, we define the energy function associated with the pair of items $(i,j)$ to be of the form $E_{ij} := \mathbf{f}_i^\top \mathbf{w} + \mathbf{f}_i^\top \mathbf{W} \mathbf{f}_j$ where $\mathbf{w} \in \R^d$ and $\mathbf{W} \in \R^{d \times d}$ are the unknown latent parameters (vector and matrix parameters repectively) to be estimated, and $\mathbf{f}_i$ and $\mathbf{f}_j$ are the known feature vectors associated with items $i$ and $j$ respectively. It is clear from Equation~\eqref{eqn:gbtl_mat} that a key advantage of the proposed model is the additional ability to incorporate side information in the form of feature vectors and feature correlations in a latent space described by $\mathbf{W}$. In matrix notation,
\begin{align}
\psi(\mathbf{P}) & = (\mathbf{1} \mathbf{g}^\top + \mathbf{F}^\top \mathbf{W}^\top \mathbf{F}) - (\mathbf{g} \mathbf{1}^\top + \mathbf{F}^\top \mathbf{W} \mathbf{F}) \nn \\
& = (\mathbf{\Sigma} \mathbf{V}^\top)^\top \mathbf{L} (\mathbf{\Sigma} \mathbf{V}^\top)
\label{eqn:psip}
\end{align}
where $\mathbf{g} := \mathbf{F}^\top \mathbf{w}$ is column vector in $\R^n$, $\mathbf{1} \in \R^n$ is the all-ones column vector, $\mathbf{F} = \mathbf{U} \mathbf{\Sigma} \mathbf{V}^\top$ is the full SVD of $\mathbf{F}$ (such that $\mathbf{U} \in \R^{d \times d}, \mathbf{V} \in \R^{n \times n}$ are orthonormal matrices with $\mathbf{\Sigma} \in \R^{d \times n}$ as the $d \times d$ diagonal matrix of singular values padded with zeros) and $\mathbf{L} := \mathbf{U}^\top ( \mathbf{1} \mathbf{w}^\top - \mathbf{w} \mathbf{1}^\top + \mathbf{W}^\top - \mathbf{W} ) \mathbf{U}$ (such that $\Sigma^{-1}_{ii} = \sigma_{i}^{-1}$ and $\Sigma^{-1}_{ij} = 0$ if $i \neq j$). It is now clear that the sufficient condition for $\mathbf{P} \in \P_n (\psi,r,\mathbf{\Sigma V}^\top)$ is that $\text{rank}(\mathbf{L}) \leq r$. Now, we describe the generality of the FLR model in Equation~\eqref{eqn:gbtl_mat} by showing that it subsumes many existing models and has much more expressiveness.
\begin{proposition}
\label{prop:lr}
The LR model is a special case of the FLR model, ie, $\P_n(\psi,r) \subseteq \P_n (\psi, r, \mathbf{A})$.
\end{proposition}

\begin{corollary}
\label{cor:lr}
Let $\mathbf{F} = \mathbf{I}$ and $\psi$ be the logit link function. From Proposition~\ref{prop:lr}, it is easy to see the following special cases from Equation~\eqref{eqn:gbtl_mat}.
\begin{enumerate}[noitemsep,nolistsep]
\item Let $\mathbf{W} = \mathbf{x} \mathbf{y}^\top$. If $\mathbf{w} = 0$, then $\mathbf{P} \in \P_n \paran{\psi,2}$. If $\mathbf{w} \neq 0$, then $\mathbf{P} \in \P_n \paran{\psi,4}$.
\item If $\mathbf{W}$ is symmetric, then $\mathbf{W} - \mathbf{W}^\top = 0$ and hence $\mathbf{P} \in \P_n \paran{\psi,2}$.
\item Let $\mathbf{\Lambda}$ be a diagonal $r \times r$ matrix; let $\{ \mathbf{X}, \mathbf{Y} \} \in \R^{n \times r}$ be orthonormal matrices. If $\mathbf{W} = \mathbf{X} \mathbf{\Lambda}_{r \times r} \mathbf{Y}^\top + \mathbf{M}$ where $\mathbf{M}$ is a symmetric matrix, then $\psi(\mathbf{P}) \in \P_n \paran{\psi,2r+2}$.
\end{enumerate}
\end{corollary}
\begin{proposition}
\label{prop:unary_ru}
The unary RU models are special cases of the FLR model, ie, $\P_n^{RU} \subseteq \P_n(\psi, r, A)$.
\end{proposition}

\begin{table*}[t]
\centering
\caption{Comparison of this work to previous works in non-active pairwise ranking: extended-BTL model is due to \protect\cite{cattelan2012models}, chest-blade model is due to \protect\cite{chen2016modeling} and low-rank model is due to \protect\cite{rajkumar2016can}. Here, $\mathbf{w}$ and $\mathbf{W}$ are model parameters as given in Equation~\eqref{eqn:gbtl_mat}, and $\mathbf{F}$ denotes the feature matrix. Furthermore, \textit{pairs} and \textit{per} denote the state-of-the-art bounds known regarding the number of pairs compared and the number of comparisons per pair respectively. Note that we consider the practically important regime of $n \gg d$.}
\label{tab:summary}
\begin{tabular}{cccccc}
\hlineB{2}
\textbf{Model} & $\mathbf{F}$ & $\mathbf{w}$ & $\mathbf{W}$ & \textit{pairs} & \textit{per} \\
\hlineB{2}
BTL & $\mathbf{F} = \mathbf{I}$ & $\mathbf{w} \in \R^n$ & $\mathbf{W} = \mathbf{0}$ & $\Omega(n \log n)$ & $\Omega(
\log n)$ \\
Item-feature & $\mathbf{F} \in \R^{d \times n}$ & $\mathbf{w} \in \R^d$ & $\mathbf{W} = \mathbf{0}$ & many & many \\
Chest-blade & $\mathbf{F} = \mathbf{I}$ & $\mathbf{w} \in \R^n$ & rank$(\mathbf{W})=O(d)$ & many & many \\
Low-rank & $\mathbf{F} = \mathbf{I}$ & $\mathbf{w} \in \R^n$ & $\mathbf{W} \in \R^{n \times n}$ & $\Omega(n r \log n)$ & $\Omega(r \log n)$ \\
\hline
\textbf{This work} & $\mathbf{F} \in \R^{d \times n}$ & $\mathbf{w} \in \R^d$ & $\mathbf{W} \in \R^{d \times d}$ & $\Omega (d^2 \log n)$ & $\Omega(d^2 \log^2 n / n^2)$ \\
\hlineB{2}
\end{tabular}
\end{table*}

\begin{corollary}
The BTL and Thurstone models are obtained as special cases of the FLR model under the logit and the probit transormations of $\mathbf{P}$ respectively. This follows from Proposition~\ref{prop:lr} (or Corollary~\ref{cor:lr}-part (1)) above together with Propositions 6 and 7 of \cite{rajkumar2016can}.
\end{corollary}

\begin{proposition}
\label{prop:if}
Regression-based models with item-specific features in \cite{cattelan2012models} are special cases of the FLR model, ie, $\P_n^{IF} \subseteq \P_n (\psi, r, \mathbf{A})$.
\end{proposition}

\begin{corollary}
Let $d \ll n$. Then we recover the blade-chest model~\cite{chen2016modeling} as a special case of the FLR model by setting $\text{rank}(\mathbf{W}) = O(d)$ and $\mathbf{w} = \mathbf{0}$. Next, when $d \geq n$, it is clear from Theorem 1 of \cite{chen2016modeling} that such preference matrices degenerate into matrices in $\P_n(\psi, n, \mathbf{A})$ where $\psi$ is the logit function. Moreover, it is easy to see that the FLR model admits both stochastic-transitive and stochastic-intransitive preference matrices.
\end{corollary}

Due to space constraints, proofs of Propositions \ref{prop:lr}, \ref{prop:unary_ru} and \ref{prop:if} are given in the appendix. To summarize, we have shown how to instantiate several previously proposed ranking models as special cases of our FLR model in Table~\ref{tab:summary}.

\section{Problem Setup and Solution Approach}
Once we have the generative ranking model as developed in the previous section, the objective in our learning problem is then to find the permutation of $n$ items that minimizes the number of violations with respect to the true underlying preference matrix $\mathbf{P}$, ie, to find the best ranking $\se$ in the sense that,
$$\se = \arg \min_{\sigma} \dist(\sigma, \mathbf{P})$$
The input is the pairwise comparison dataset $S = \{(i,j,y_{ij}^k)\}$ which consists of comparison results of pairs $(i,j)$ from a survey involving $K$ users where each user with index $k$ assigns $y_{ij}^k = 1$ if he prefers $i$ to $j$ and $y_{ij}^k = 0$ if he prefers $j$ to $i$. Note that it is not necessary that all pairs of items be compared; our algorithm is able to handle noisy and incomplete data. Since the true preference matrix $\mathbf{P}$ is unknown, our algorithm instead proceeds by using the empirical preference matrix $\mathbf{\pe}$ computed from the available $y_{ij}^k$; it is to be noted, even then, our analysis guarantees that $\dist(\se, \mathbf{P})$ is good as opposed to just $\dist(\se, \mathbf{\pe})$. Additionally, in our inductive setting, the feature information is encoded by $\mathbf{f}_i \in \R^d$ for every item $i$ and concatenated to form the feature matrix $\mathbf{F} \in \R^{d \times n}$.

\subsection{Algorithm}
We present our main algorithm for inductive ranking in Algorithm~\ref{alg:ipr}. The input data consist of the set of pairwise comparison results $S = \{(i, j, \{ y_{ij}^k \})\}$, $(i,j) \in \Xi \subseteq [n] \times [n]$, $k \in [K]$, $y_{ij}^k \in \{0,1\}$ and the feature matrix $\mathbf{F} \in \R^{d \times n}$. The algorithm assumes the link function and the rank as input parameters. The subroutines used are:
\begin{enumerate}
\item \textit{Noisy matrix completion with features (Subroutine~\ref{alg:imc}): }Note that to solve our ranking problem and derive the associated recovery guarantee, it suffices, as we have done, to use the specified trace-norm program as a black-box method; hence, we assume that we have access to an oracle that gives us the solution to the convex program. The details of how the solution to this program may be found numerically is beyond the scope of this work -- for further details regarding some possible sub-gradient algorithms, we refer the reader to \cite{chiang2015matrix} and \cite{ji2009accelerated}.
\item \textit{$\gamma$-approximate pairwise ranking procedure (Subroutine~\ref{alg:pr}): }Let  $\se \in \S_n$ be the output of any Pairwise Ranking (PR) procedure with respect to an underlying preference matrix $\mathbf{P}$. For a constant $\gamma > 1$, $\se$ is said to be $\gamma$-approximate if $\dist(\se,\mathbf{P}) \leq \gamma \min_{\sigma \in \S_n} \dist(\sigma,\mathbf{P})$. Any constant factor approximate ranking procedure maybe used. Specifically, we use the Copeland procedure~\cite{copeland1951reasonable} as a black-box method which has a $5$-approximation guarantee~\cite{coppersmith2006ordering}. This method involves simply sorting the items according to a score which is computed for every item $i$ as $\sum_{j = 1}^n \mathbbm{1} (\pec_{ij} > 1/2)$.
\end{enumerate}


\section{Analysis}
In this section, we state and prove our main result.
\begin{theorem}[\textbf{Guaranteed rank aggregation with sub-linear sample complexity using item features}]
\label{thm:main}
Let $\mathbf{P} \in (\P_n (\psi,r,\mathbf{A}) \cap \P_n^{ST})$ be the true underlying preference matrix according to which the pairwise comparison dataset $S = \{(i, j, \{ y_{ij}^k \})\}$ is generated. Let $\psi$ be $L$-Lipschitz in $[\frac{P_{\min}}{2},1-\frac{P_{\min}}{2}]$. Let $\Xi$ be the set of pairs of items compared such that the number of pairs compared is $\abs{\Xi} = m > \frac{48 C_2^2 d^2 \log(n) (1+\gamma)^2}{\kappa^8 \epsilon^2 \Delta^4}$ where $\Xi$ is chosen uniformly at random from among all possible subsets of item pairs of size $m$. Let each pair in $\Xi$ be compared independently $K \geq \frac{16 (1+\gamma) m L^2 \log(n)}{n^2 \Delta^2 \epsilon}$ times where $\Delta = \min_{i \neq j} \abs{\psi(P_{ij})-\psi(1/2)}$. Then, with probability atleast $1-3/n^3$, for any $\epsilon > 0$, Algorithm~\ref{alg:ipr} returns an estimated permutation $\se$ such that $\dist(\se, \mathbf{P}) \leq \epsilon$.
\end{theorem}
\begin{remark}
The key take-away message in Theorem~\ref{thm:main} is the reduction in sample complexity possible due to efficient utilization of features and feature correlations, associated with the items to be ranked, by Algorithm~\ref{alg:ipr}. 
For instance, when $d = O(1)$, which is often the case in practice, we reduce the required total number of comparisons to be made to $\Omega(\log (n))$. Thus, we achieve a very significant gain since the total number of comparisons is poly-logarithmic as opposed to quadratic in the number of items. This is especially crucial in large-scale machine learning applications.
\end{remark}
\begin{remark}
Another point to be noted from Theorem~\ref{thm:main} is that, under the uniform sampling assumption, when features associated with items are known, it is more important that we compare sufficient (precisely, $\Omega(\log n)$) number of different pairs rather than high number of comparisons per pair. Furthermore, he total number of comparisons needed in Theorem~\ref{thm:main} is given by the product $m K$ which is $\Omega(\max(\log n / \epsilon^2, d^4 \log^3 n / \epsilon^3 n^2))$.
\end{remark}
\noindent We now present the proof of Theorem~\ref{thm:main}. We shall prove the theorem under the Bernoulli sampling model (where each entry of an $n \times n$ matrix is observed independently with probability $1/n^2$) rather than the uniform sampling model (wherein $\Xi$ is chosen uniformly at random from among all possible subsets of item pairs of size $m$); the equivalence between the two is well-known (see, for instance, Section 7.1 of \cite{candes2011robust}).

\floatname{algorithm}{Subroutine}
\begin{algorithm}[t]
\caption{IMC: Inductive Matrix Completion}
\label{alg:imc}
\begin{algorithmic}[1]
\REQUIRE $M_{ij}$ for $(i,j) \in \Xi \subseteq [n] \times [n]$, feature matrix $\mathbf{F}$.
\ENSURE Completed matrix $\overline{\mathbf{M}}$.
\STATE Solve the convex program:
\begin{align*}
\widehat{\mathbf{Z}} = \arg \min_{\mathbf{Z_L}} & \frobnorm{\proj_\Xi ( \mathbf{M} - \mathbf{F}^\top \mathbf{Z_L F} ) }^2 \text{ s.t. } \trnorm{\mathbf{Z_L}} \leq C_L
\end{align*}
\RETURN $\psi(\mathbf{\pec}) \leftarrow \mathbf{F}^\top \widehat{\mathbf{Z}} \mathbf{F}$.
\end{algorithmic}
\end{algorithm}
\floatname{algorithm}{Subroutine}
\begin{algorithm}[t]
\caption{PR: Pairwise Ranking (Copeland Procedure)}
\label{alg:pr}
\begin{algorithmic}[1]
\REQUIRE Preference matrix $\mathbf{M} \in \R^{n \times n}$.
\ENSURE Ranking $\se$.
\STATE Threshold: $\forall (i,j), \quad \widetilde{M}_{ij} \leftarrow \mathbbm{1}(M_{ij} > 1/2)$.
\STATE Compute row-sum of $\mathbf{\widetilde{M}}$: $\mathbf{v} \leftarrow \mathbf{\widetilde{M}} \mathbf{1}$.
\RETURN $\se \leftarrow \text{Sort}(\mathbf{v})$.
\end{algorithmic}
\end{algorithm}

\floatname{algorithm}{Algorithm}
\begin{algorithm}[t]
\caption{IPR: Inductive Pairwise Ranking}
\label{alg:ipr}
\begin{algorithmic}[1]
\REQUIRE Set of comparison results $S = \{(i, j, \{ y_{ij}^k \})\}$, feature matrix $\mathbf{F}$, link function $\psi$, target rank $r$.
\ENSURE Ranking of $n$ items, $\hat{\sigma} \in \S_n$.
\STATE Construct the partially observed empirical preference matrix using $S$ as:
\[
\pe_{ij} =
\begin{cases}
\frac{1}{K} \sum_{k=1}^K y_{ij}^k & \text{ if } (i,j) \in \Xi \\
\frac{1}{K} \sum_{k=1}^K (1-y_{ij}^k) & \text{ if } (j,i) \in \Xi \\
1/2 & \text{ if } i=j \text{ or } (i,j) \notin \Xi \\
\end{cases}
\]
\STATE Compute SVD of $\mathbf{F} = \mathbf{U \Sigma V}^\top$ and set $\mathbf{A} \leftarrow \mathbf{\Sigma V}^\top$.
\STATE Use a noisy inductive matrix completion subroutine: $\mathbf{\psi(\pec)} \leftarrow \text{IMC}(\mathbf{\psi(\pe)}, \mathbf{A})$.
\STATE Take the inverse transform of the truncated $r$-SVD of the completed matrix estimate: $\mathbf{Q} \leftarrow \psi^{-1} ( \P_r (\psi(\mathbf{\pec})) )$.
\STATE Using a pairwise ranking subroutine: $\se \leftarrow \text{PR}(\mathbf{Q})$.
\RETURN $\se$.
\end{algorithmic}
\end{algorithm}

\begin{proof}
Let $\pe_{ij}$ be the empirical probability estimate of $P_{ij}$. Note that we compute  $\pe_{ij} = \frac{1}{K} \sum_{k=1}^K y_{ij}^k$ for $(i,j) \in \Xi$ from the given pairwise comparison dataset, $S = \{(i, j, \{ y_{ij}^k \})\}$. From Equation~\eqref{eqn:psip}, $\psi(\mathbf{P}) = \mathbf{A}^\top \mathbf{L} \mathbf{A}$ where $\mathbf{A} = \mathbf{\Sigma V}^\top$. Since we use the empirical estimate for $P_{ij}$, we have noise due to sampling error only over $\Xi$, ie, $\psi(\mathbf{\pe}) = \psi(\mathbf{P}) + \mathbf{N} = \mathbf{A}^\top \mathbf{L A} + \mathbf{N}$ where
\begin{align*}
\abs{N_{ij}} =
\begin{cases}
0 & \text{ if } (i,j) \notin \Xi \\
\abs{\psi( \frac{1}{K} \sum_{k=1}^K y_{ij}^k) - \psi(P_{ij})} & \text{ if } (i,j) \in \Xi
\end{cases}
\end{align*}
Now, we solve the trace-norm regularized convex program corresponding to the noisy inductive matrix completion problem:
\begin{align*}
\{ \overline{\mathbf{L}}, \overline{\mathbf{N}} \} = \arg \min_{\mathbf{Z_N}, \mathbf{Z_L}} & \frobnorm{\proj_\Xi ( \psi(\mathbf{\pe}) - (\mathbf{A}^\top \mathbf{Z_L} \mathbf{A} + \mathbf{Z_N}) ) }^2 \\
& + \lambda_L \trnorm{\mathbf{Z_L}} + \lambda_N \trnorm{\mathbf{Z_N}}
\end{align*}
and let $\psi(\mathbf{\pec}) = \mathbf{A}^\top \overline{\mathbf{L}} \mathbf{A} + \overline{\mathbf{N}}$ be the link-transformed completed (estimate) matrix where $\overline{\mathbf{N}}$ be the estimated noise matrix. This is equivalent to solving the problem:
\begin{align*}
\{ \overline{\mathbf{L}}, \overline{\mathbf{N}} \} = \arg \min_{\mathbf{Z_N}, \mathbf{Z_L}} & \frobnorm{\proj_\Xi ( \psi(\mathbf{\pe}) - (\mathbf{A}^\top \mathbf{Z_L} \mathbf{A} + \mathbf{Z_N}) ) }^2 \\
& \text{ s.t. } \trnorm{\mathbf{Z_L}} \leq C_L , \quad  \trnorm{\mathbf{Z_N}} \leq C_N
\end{align*}
We set $C_N = 0$ and $C_L = \trnorm{ (\mathbf{A}^\top)^\dagger \psi(\mathbf{\pe}) (\mathbf{A})^\dagger }$ which may be upper bounded, by Lemma 3 of \cite{chiang2015matrix} as $C_L \leq \frac{d}{C' \kappa^4}$ for a constant $C'$. We now recall Theorem 1 from \cite{chiang2015matrix}. Let $\delta < 1/d$ and $\mathbf{A}$ be well-conditioned, specifically, $\kappa^4 \leq C_2 d$ for constant $C_2$. The expected squared loss under Bernoulli sampling is bounded as, with probability at least $1-\delta$: 
\begin{align}
\frac{\frobnorm{ \psi(\mathbf{\pec}) - \psi(\mathbf{\pe}) }^2}{n^2} & \leq C_1 \min \paran{ C_N \sqrt{\frac{\log(2n)}{m}}, \sqrt{C_N \frac{\sqrt{n}}{m}} } \nn \\
& + \frac{C_2 d}{\kappa^4} \sqrt{\frac{\log (2/\delta)}{m}}
\label{eqn:risk}
\end{align}
where $C_1$ and $C_2$ are constants. By triangle inequality,
\begin{align*}
\frobnorm{ \psi(\mathbf{\pec}) - \psi(\mathbf{\pe}) } & = \frobnorm{ \psi(\mathbf{\pec}) - (\psi(\mathbf{P})+\mathbf{N}) } \\
& \geq \frobnorm{ \psi(\mathbf{\pec}) - \psi(\mathbf{P}) } - \frobnorm{\mathbf{N}}
\end{align*}
Using $C_N = 0$ in Equation~\eqref{eqn:risk}, with probability at least $1-\delta$,
\begin{align*}
\frac{1}{n} \frobnorm{ \psi(\mathbf{\pec}) - \psi(\mathbf{P}) } \leq \paran{ \frac{C_2 d}{\kappa^4} \sqrt{\frac{\log (2/\delta)}{m}} }^{1/2} + \frac{1}{n} \frobnorm{\mathbf{N}}
\end{align*}
Let $K \geq \frac{m L^2 \log(n)}{\tau^2}$ where $\tau = n \sqrt{\frac{\epsilon}{1+\gamma}} \frac{\Delta}{4}$. Substituting the bounds for the $\mathbf{N}$ terms from Lemma~\ref{lem:sampling_noise} and using the union bound, with probability at least $1-\delta - 1/n^3$,
\begin{align*}
& \frobnorm{ \psi(\mathbf{\pec}) - \psi(\mathbf{P}) } \leq n \paran{ \frac{C_2 d}{\kappa^4} \sqrt{\frac{\log (2/\delta)}{m}} }^{\frac{1}{2}} + \tau \\
& \leq n \paran{ \frac{C_2 d}{\kappa^4} \sqrt{\frac{\log (2/\delta)}{m}} }^{\frac{1}{2}} + n \sqrt{\frac{\epsilon}{1+\gamma}} \frac{\Delta}{4}
\end{align*}
Now, setting $m > \frac{16 C_2^2 d^2 \log(2/\delta) (1+\gamma)^2}{\kappa^8 \epsilon^2 \Delta^4}$ and $\delta = 2/n^3$, we obtain, with probability $1-3/n^3$,
$\frobnorm{ \psi(\mathbf{\pec}) - \psi(\mathbf{P}) } \leq 
n \sqrt{\frac{\epsilon}{1+\gamma}} \frac{\Delta}{2}$. 
Then, arguments similar to the proof of Theorem 13 of \cite{rajkumar2016can} 
 yield our result.
\end{proof}
\begin{lemma}[\textbf{Characterization of noise due to finite-sample effects}]
\label{lem:sampling_noise}
Under the conditions of Theorem~\ref{thm:main}, let $m$ item pairs be compared such that the number of comparisons per item pair is $K \geq \frac{m L^2 \log(n)}{\tau^2}$. Then, with probability atleast $1- 1 / n^3$, $\frobnorm{\mathbf{N}} \leq \tau$.
\end{lemma}
Due to space limitations, the proof of Lemma~\ref{lem:sampling_noise} is given in the appendix.

\begin{figure}[t]
\caption{Ranking results of LRPR and IPR: fixing $d = 20$ and $K = 50 \lceil d^2 \log^2(n) / n^2 \rceil$ while varying $m$.}
\label{fig:vary_m}
\includegraphics[width = 0.3\columnwidth]{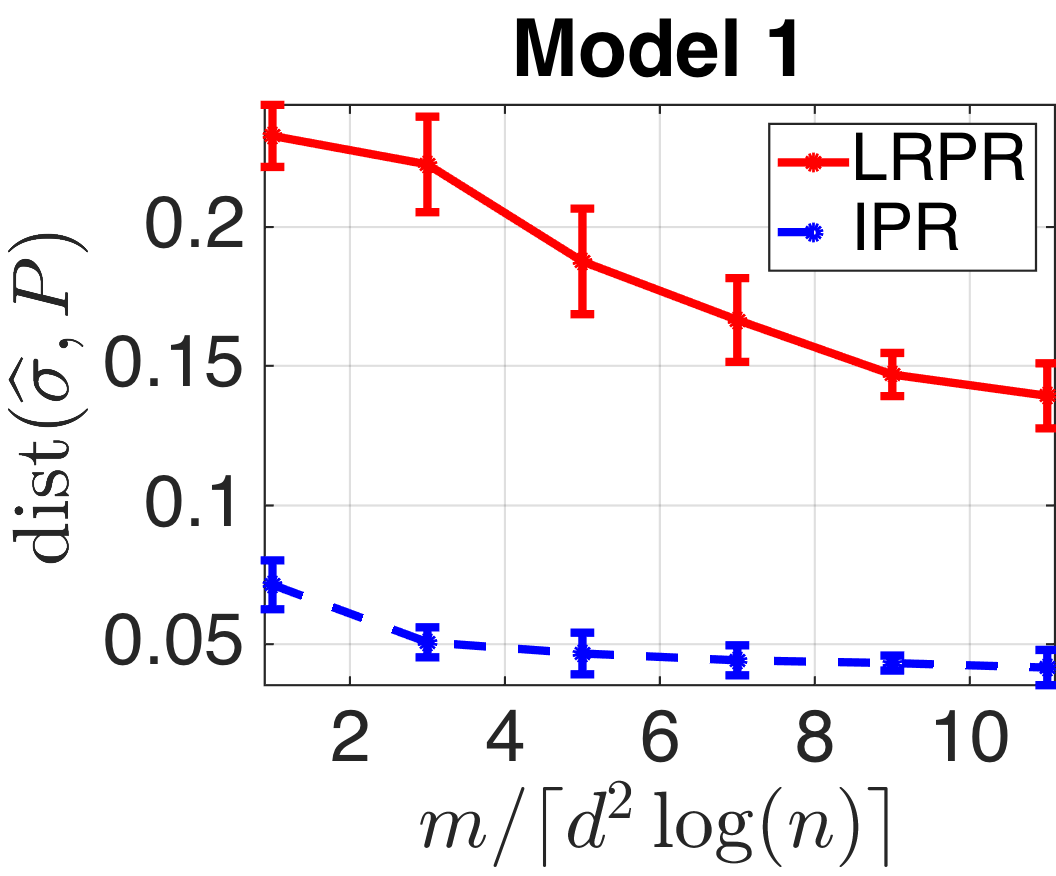}
\includegraphics[width = 0.3\columnwidth]{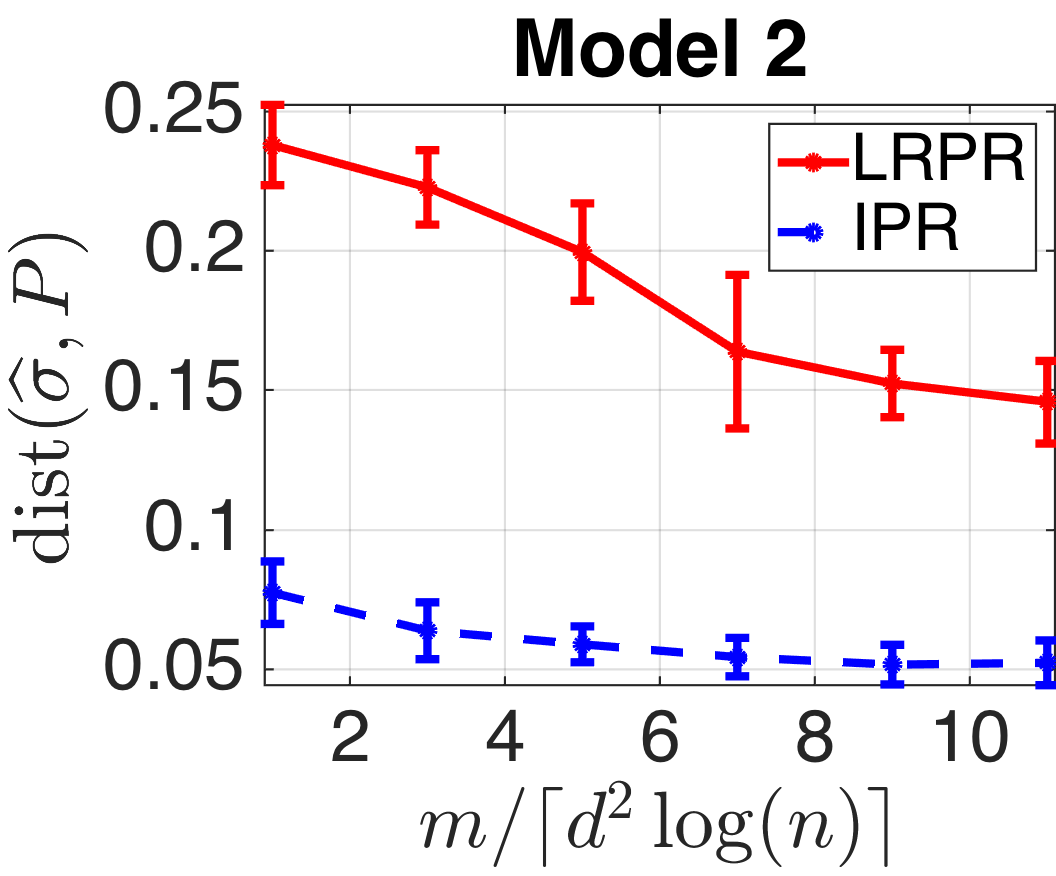}
\includegraphics[width = 0.3\columnwidth]{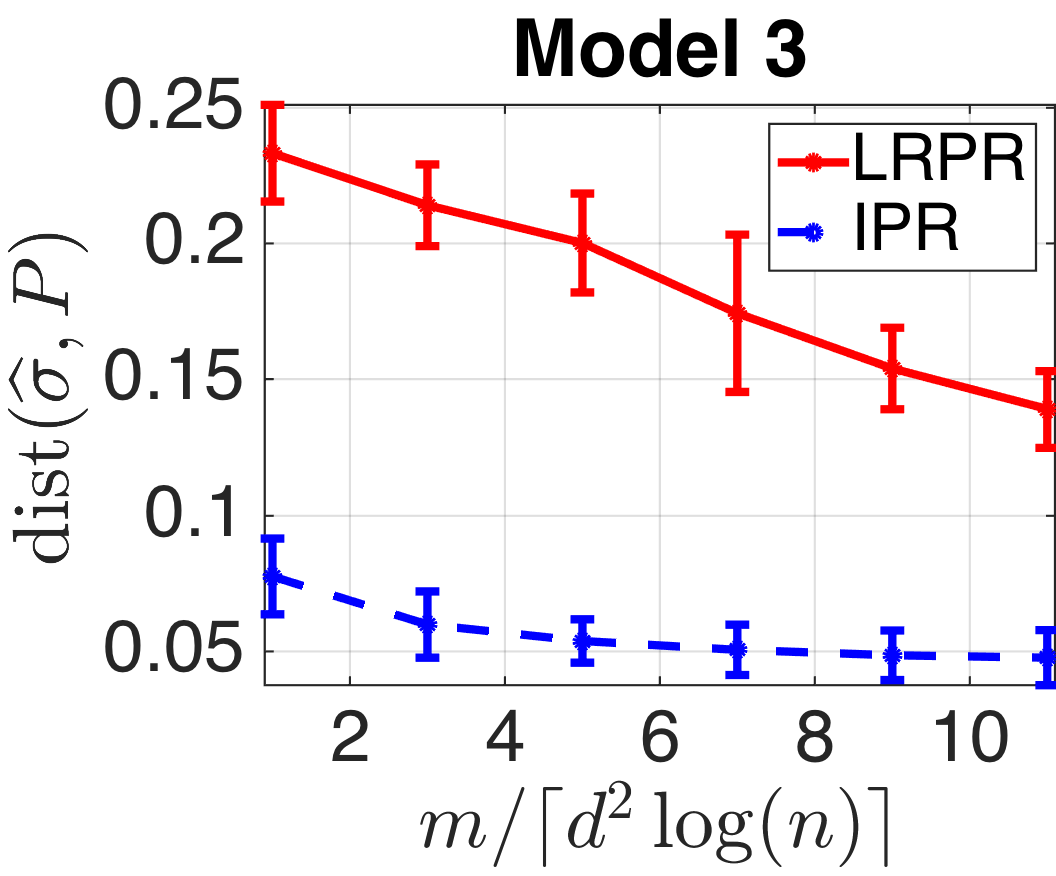}
\end{figure}
\begin{figure}[t]
\caption{Ranking results of LRPR and IPR: fixing $d = 20$ and $m = \lceil d^2 \log(n) \rceil$ while varying $K$.}
\label{fig:vary_K}
\includegraphics[width = 0.3\columnwidth]{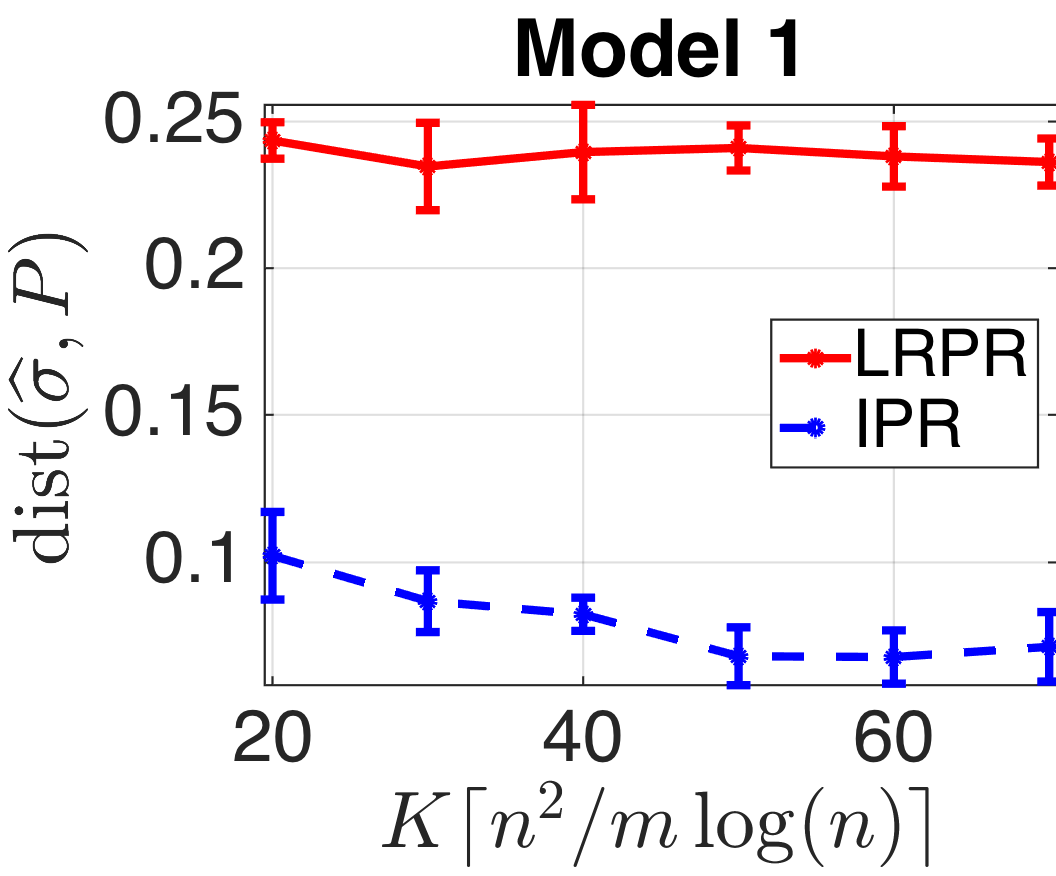}
\includegraphics[width = 0.3\columnwidth]{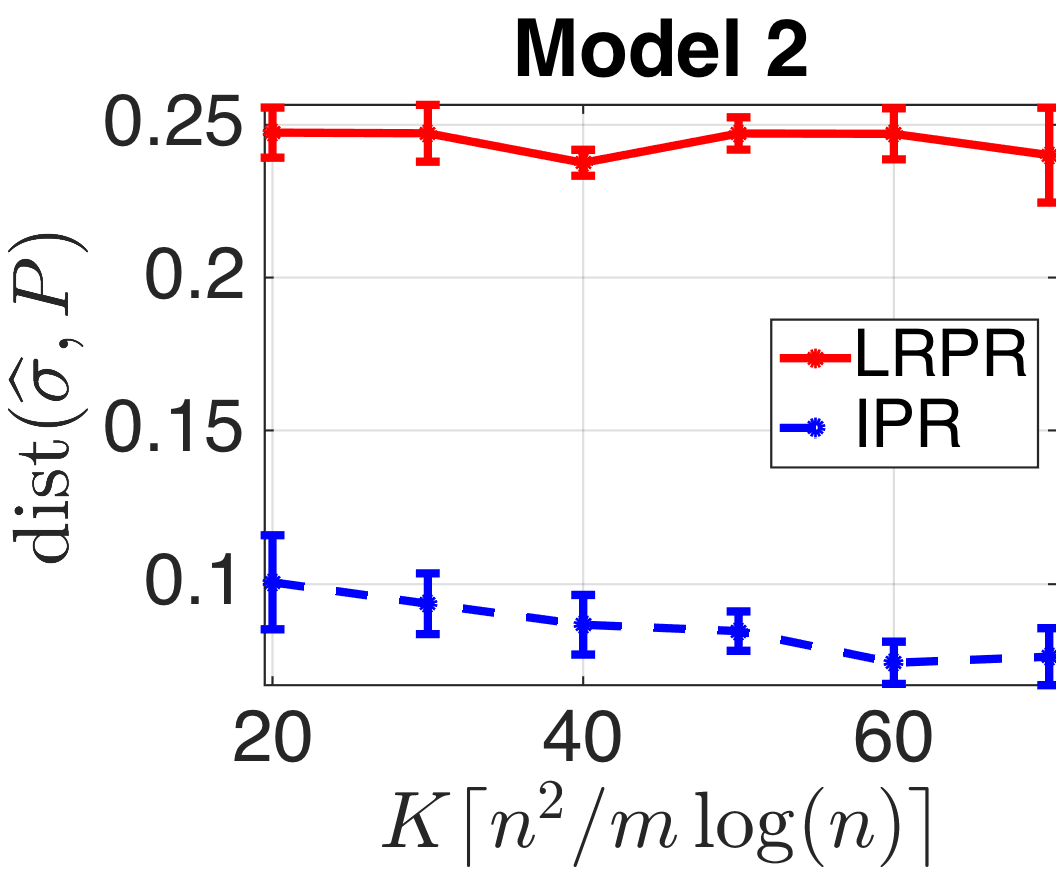}
\includegraphics[width = 0.3\columnwidth]{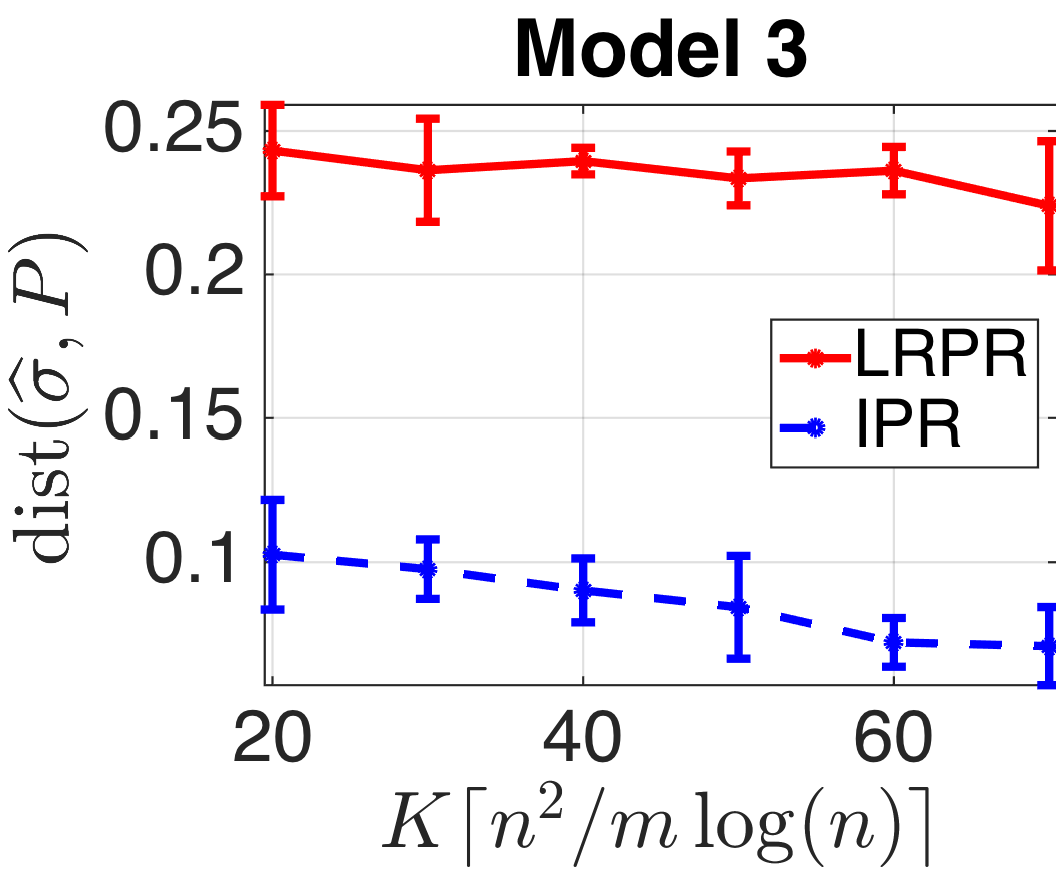}
\end{figure}
\begin{figure}[t]
\caption{Ranking results of LRPR and IPR: fixing $K = 50 \lceil d^2 \log^2(n) / n^2 \rceil$ and $m = \lceil d^2 \log(n) \rceil$ while varying $d$.}
\label{fig:vary_d}
\includegraphics[width = 0.3\columnwidth]{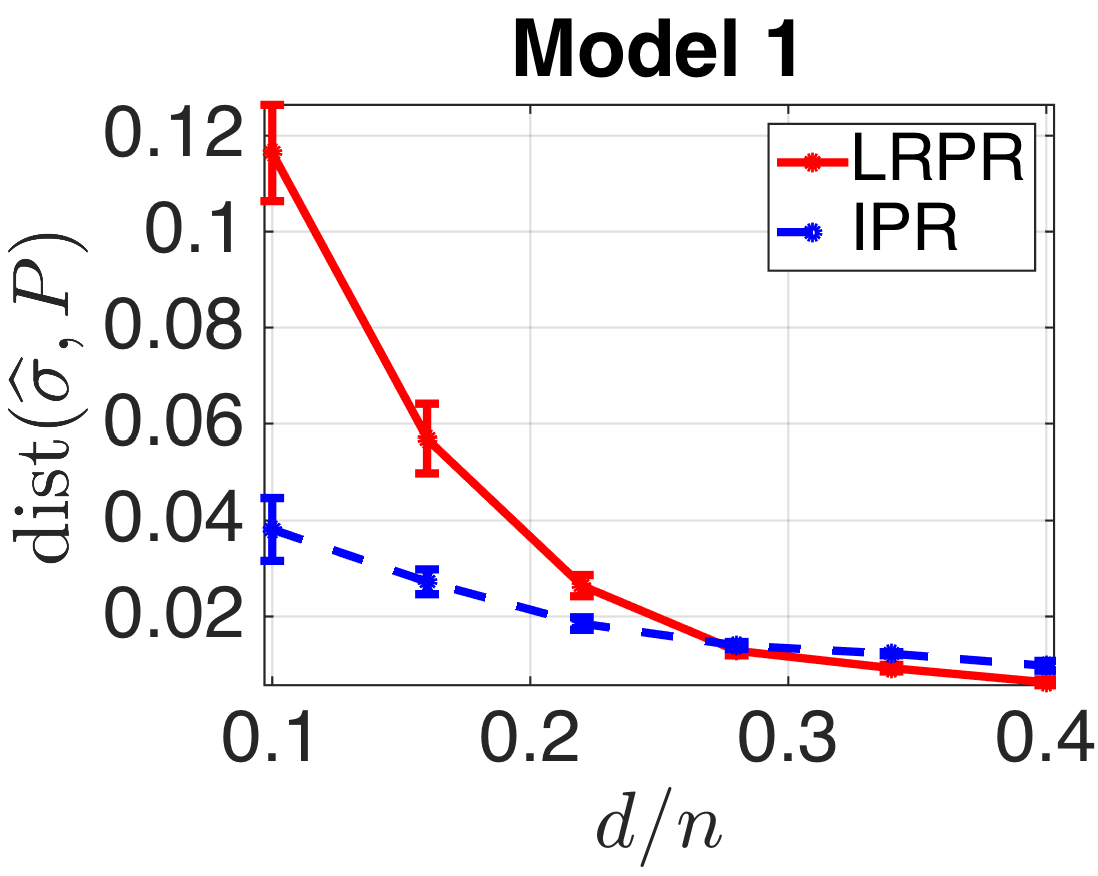}
\includegraphics[width = 0.3\columnwidth]{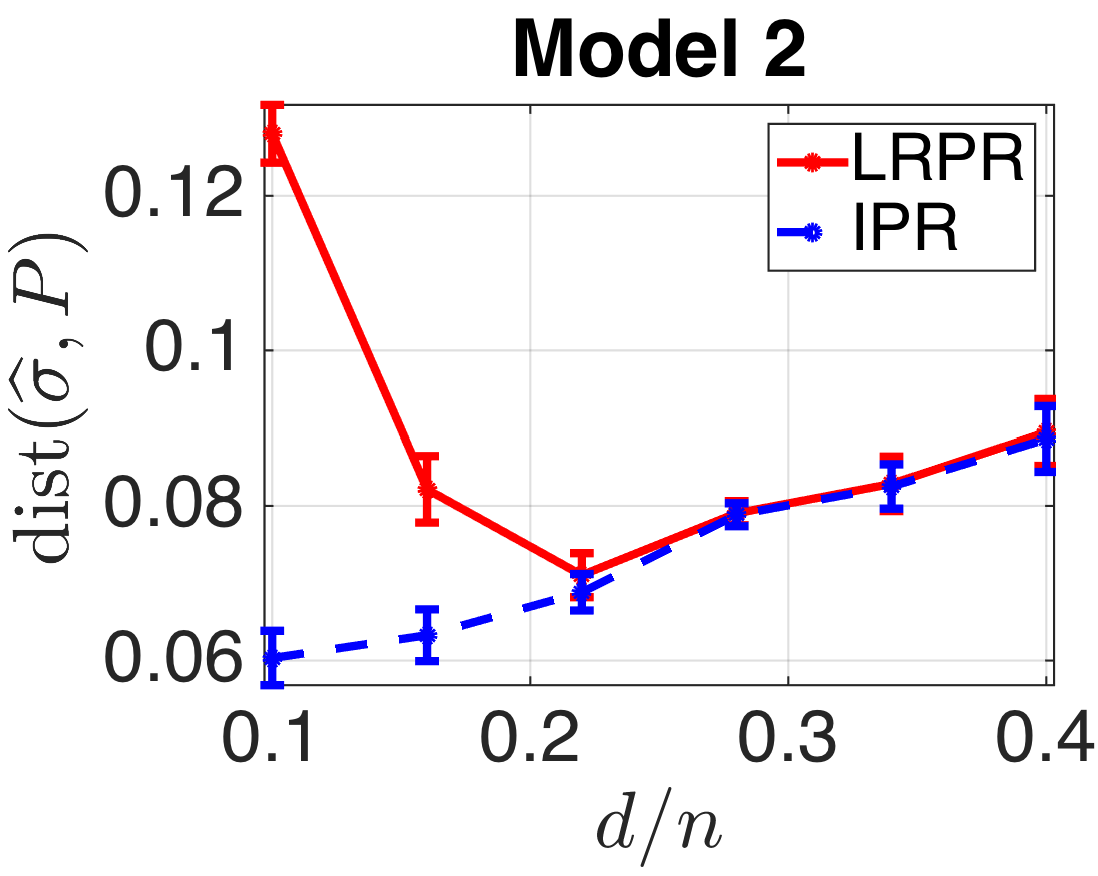}
\includegraphics[width = 0.3\columnwidth]{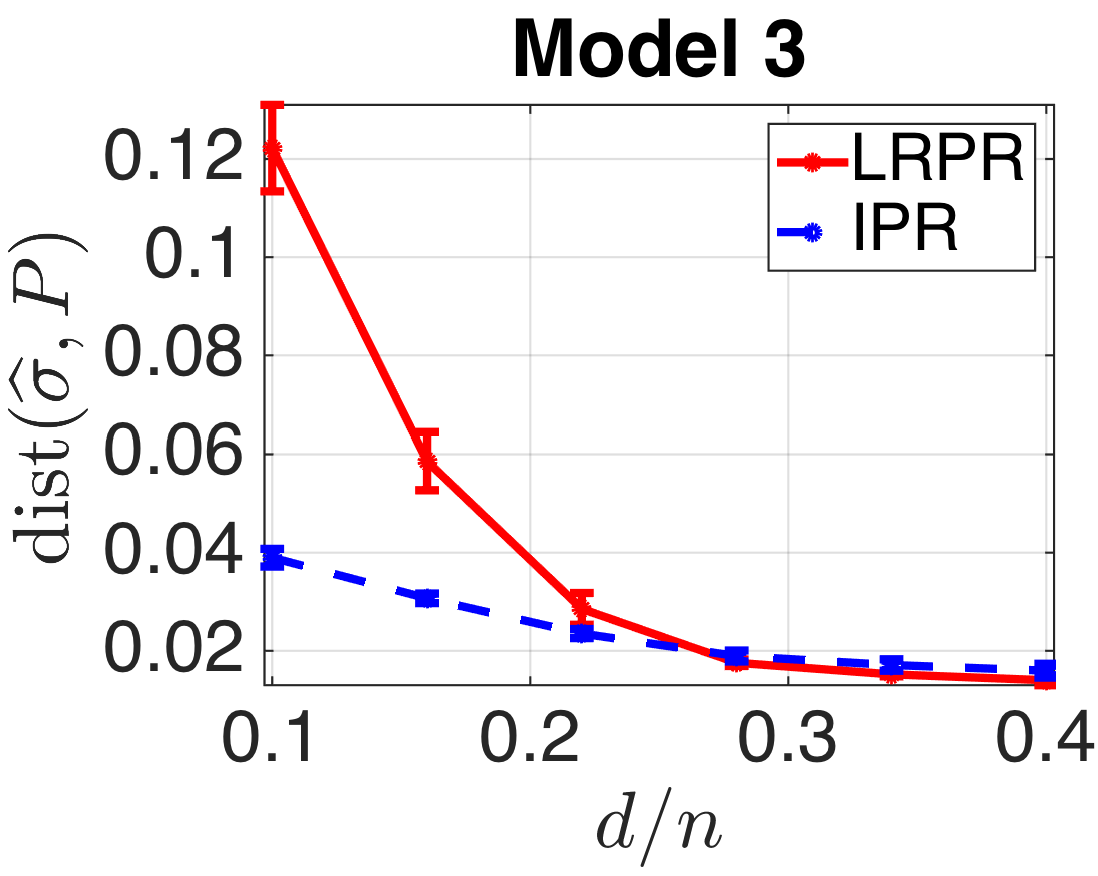}
\end{figure}
\begin{figure}[t]
\caption{Ranking results of LRPR and IPR on Sushi and Car datasets: we fix $m = \lceil d^2 \log(n) \rceil$ while varying $K$; we fix $K = 50 \lceil d^2 \log^2(n) / n^2 \rceil$ while varying $m$}
\label{fig:vary_real}
\includegraphics[width = 0.24\columnwidth]{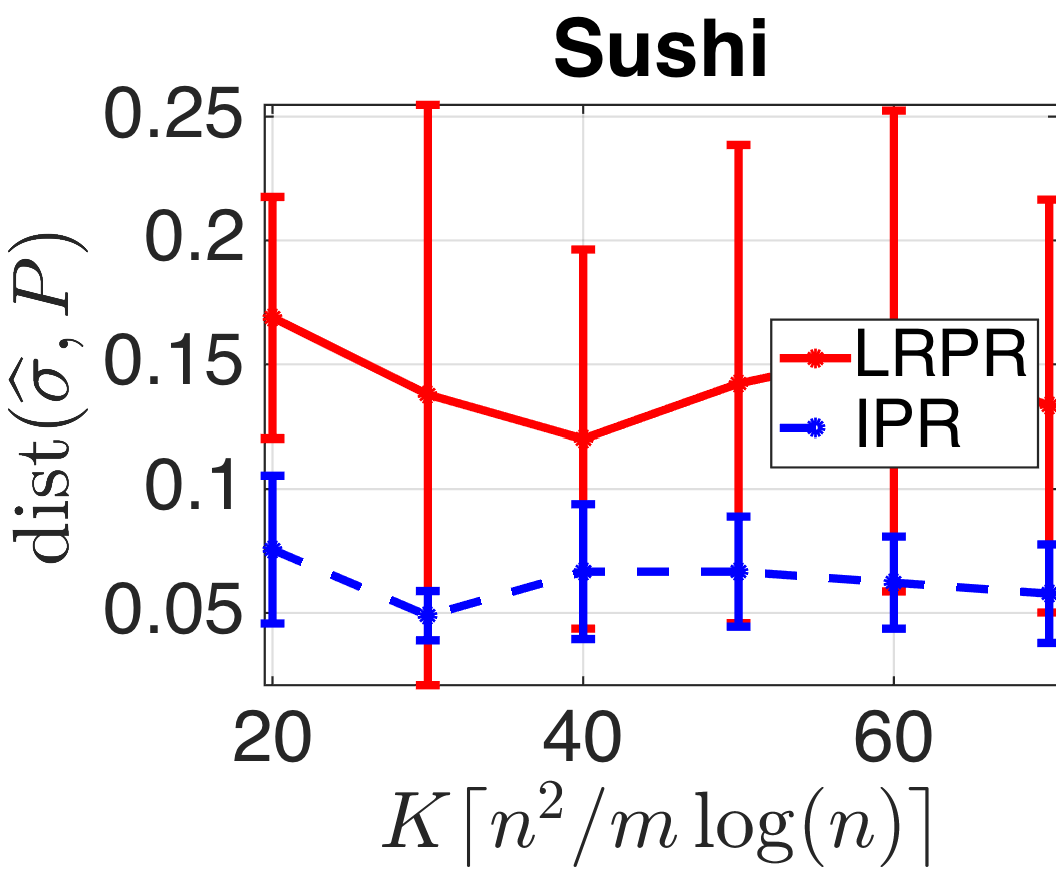}
\includegraphics[width = 0.24\columnwidth]{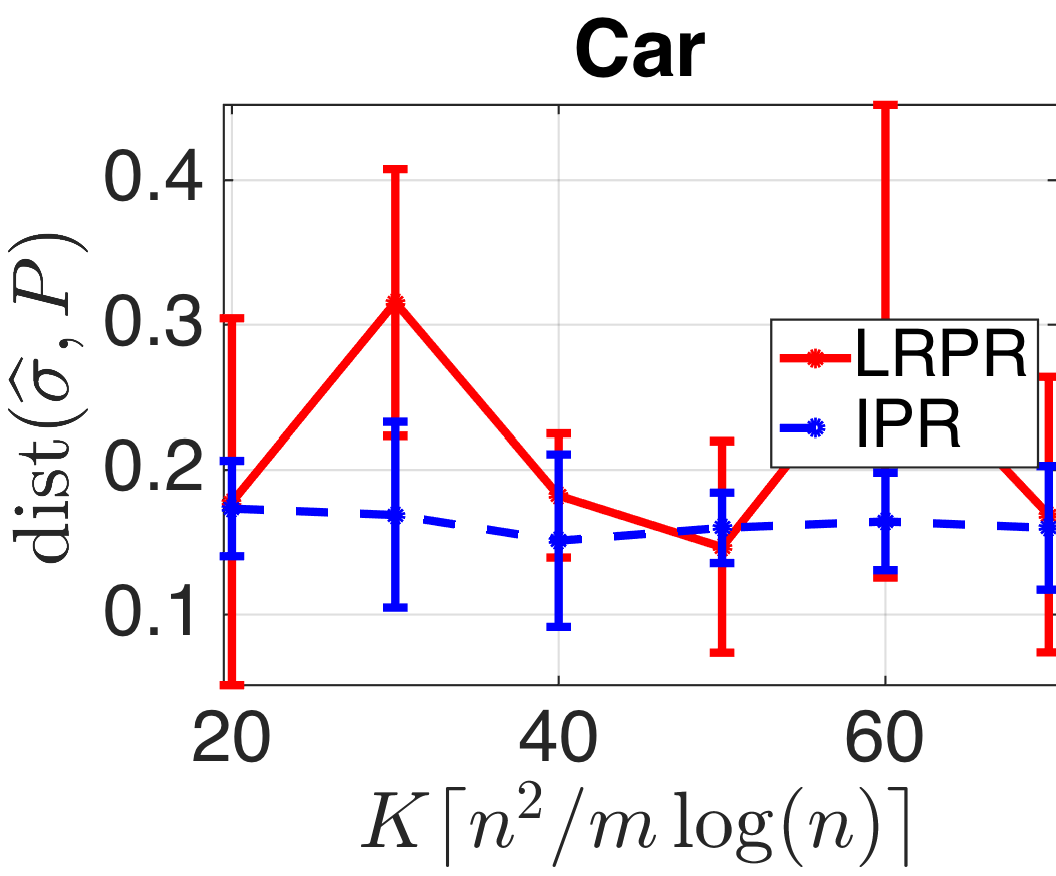}
\includegraphics[width = 0.24\columnwidth]{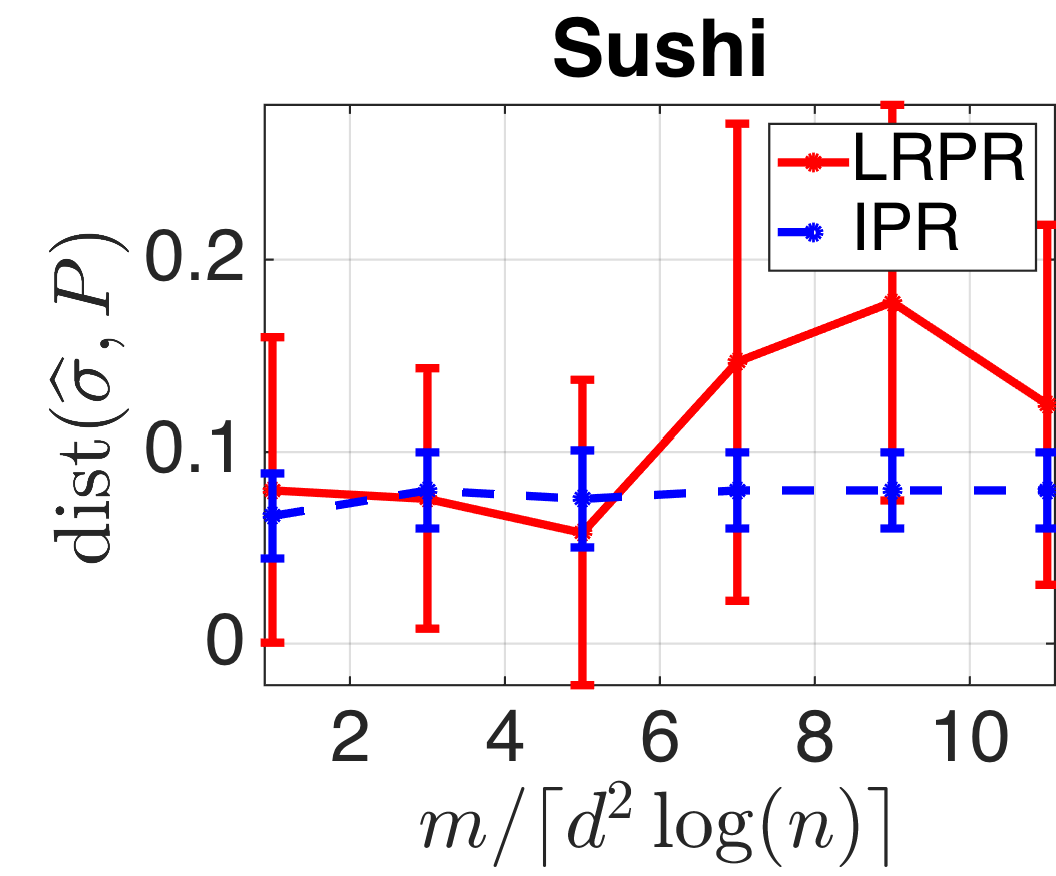}
\includegraphics[width = 0.24\columnwidth]{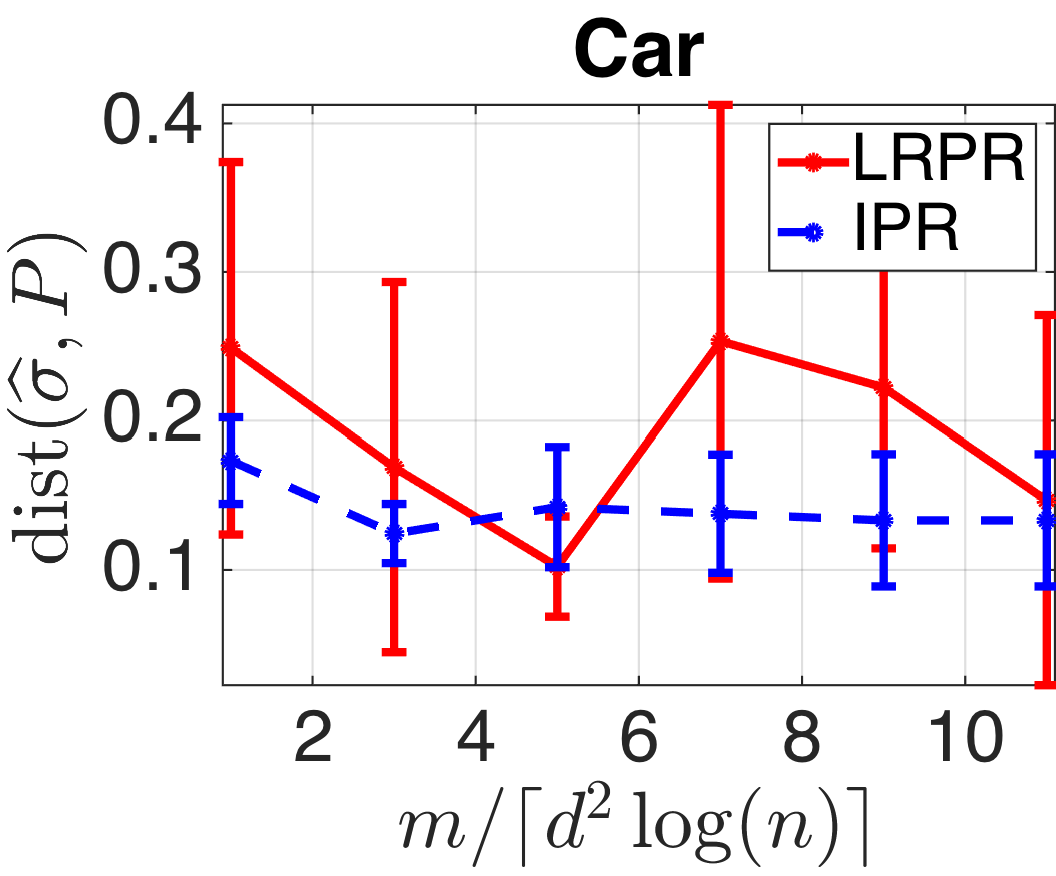}
\end{figure}

\section{Experimental Results}
In this section, we conduct a systematic empirical investigation of the performance of our ranking method and justify our theoretical claim in the previous section. The goal of this study is two-fold: (a) to verify the correctness of our algorithm, and (b) to show that by using features and feature correlations, our IPR algorithm has a better sample complexity thereby improving upon the LRPR algorithm that does not take into account the available side information.

\subsection{Synthetic Simulations}
For a given set of $n = 500$ items, we consider three main problem parameters:
(1) $m$ -- the number of item pairs compared (Figure~\ref{fig:vary_m}),
(2) $K$ -- the number of comparisons per pair (Figure~\ref{fig:vary_K}),
(3) $d$ -- the dimensionality of features (Figure~\ref{fig:vary_d}).
We study the performance of both IPR and LRPR algorithms by varying each of the problem parameters while fixing the others. We note that by making use of side information, IPR outperforms LRPR in all the cases as shown in the sample complexity plots. 
All the accuracy results presented are obtained by averaging over five runs.
\paragraph{Data generation: }We consider three representative preference matrices derived from Equation~\eqref{eqn:gbtl_mat}: (a) \textit{Model-1}: we set $W = 0$, (b) \textit{Model-2}: we construct a general $W$; here, we generate $W_{ij} {\sim} U(0,1)$, and (c) \textit{Model-3}: we construct a low-rank $\mathbf{W}$, ie, $\text{rank}(\mathbf{W}) = 2 < d$; here we generate $W_{ij} {\sim} U(0,1)$ and then truncating $\mathbf{W}$ by setting all but its top two singular values to zero. In all the three models, we generate $w_i {\sim} U(0,1)$. The features are generate as $F_{ij}{\sim} U(0,1)$; to ensure that the features are well-conditioned, we perform the full SVD of feature matrix $\mathbf{F}$ and set all its singular values to 1.
\paragraph{Parameter settings: }For IPR, we choose $\lambda_L = 10^{-2}$ and $\lambda_N = 10^{2}$. Note that LRPR allows for the rank of the problem to be automatically determined. In the same spirit, though Step-5 of Algorithm~\ref{alg:ipr} requires the knowledge of the true rank, we choose \textit{not} to perform this truncation step thereby including the error induced by the smaller singular values resulting from noise due to sampling in our distance estimate -- even then, IPR outperforms LRPR.

\subsection{Real-data Simulations}
We apply our method on two popular preference learning datasets. We briefly describe the data and the results (Figure~\ref{fig:vary_real}) we obtain below:
\begin{enumerate}
\item \textit{Sushi: }This data~\cite{kamishima2009efficient} is from a survey of $5000$ customers. Each customer orders $10$ sushi dishes according to their preferences. The goal, then, is to estimate a global ranking of these sushi dishes using these observations from customers. Each sushi has six features such as price, taste and so on. We construct the complete preference matrix $\mathbf{P} \in [0,1]^{10 \times 10}$ using the preferences of all the customers and consider this to be ground truth preference matrix. An interesting observation was that, over five runs of the algorithms, IPR gets two out of the top four sushi dishes right most of the times namely, `amaebi' and `ikura'; on the other hand, LRPR does not succeed in recovering these always. 
\item \textit{Car: }The task in this dataset~\cite{abbasnejad2013learning} is find an order of preference among ten cars. This data was collected by surveying $60$ customers regarding there preferences among pairs of cars drawn from the set of ten cars. Each car has four features including engine, transmission and so on. We construct the ground truth preference matrix $\mathbf{P} \in [0,1]^{10 \times 10}$ by aggregating the pairwise preferences of all the customers. An interesting trend we found was that customers generally preferred sedans over SUVs and non-hybrid vehicles over hybrid vehicles.
\end{enumerate}


\section{Discussion and Future Directions}
In this paper, we have proposed and characterized the FLR model together with the guaranteed IPR algorithm that utilizes available side information of the items to be ranked to provably reduce the sample complexity for ranking from $\Omega (n \log n)$ to possibly as low as $\Omega (\log n)$. A future research direction is to see if mixture models for ranking such as the recently proposed topic modeling approach~\cite{ding2015topic} could fit into our framework while admitting sample-efficient estimation algorithms.



\begin{thebibliography}{}

\bibitem[\protect\citeauthoryear{Abbasnejad \bgroup et al\mbox.\egroup
  }{2013}]{abbasnejad2013learning}
Abbasnejad, E.; Sanner, S.; Bonilla, E.~V.; Poupart, P.; et~al.
\newblock 2013.
\newblock Learning community-based preferences via dirichlet process mixtures
  of gaussian processes.
\newblock In {\em IJCAI}.

\bibitem[\protect\citeauthoryear{Bradley and Terry}{1952}]{bradley1952rank}
Bradley, R.~A., and Terry, M.~E.
\newblock 1952.
\newblock Rank analysis of incomplete block designs: I. the method of paired
  comparisons.
\newblock {\em Biometrika} 39(3/4):324--345.

\bibitem[\protect\citeauthoryear{Cand{\`e}s and Recht}{2009}]{candes2009exact}
Cand{\`e}s, E.~J., and Recht, B.
\newblock 2009.
\newblock Exact matrix completion via convex optimization.
\newblock {\em Foundations of Computational mathematics} 9(6):717--772.

\bibitem[\protect\citeauthoryear{Cand{\`e}s \bgroup et al\mbox.\egroup
  }{2011}]{candes2011robust}
Cand{\`e}s, E.~J.; Li, X.; Ma, Y.; and Wright, J.
\newblock 2011.
\newblock Robust principal component analysis?
\newblock {\em Journal of the ACM (JACM)} 58(3):11.

\bibitem[\protect\citeauthoryear{Caragiannis, Procaccia, and
  Shah}{2013}]{caragiannis2013noisy}
Caragiannis, I.; Procaccia, A.~D.; and Shah, N.
\newblock 2013.
\newblock When do noisy votes reveal the truth?
\newblock In {\em Proceedings of the fourteenth ACM conference on Electronic
  commerce},  143--160.
\newblock ACM.

\bibitem[\protect\citeauthoryear{Cattelan}{2012}]{cattelan2012models}
Cattelan, M.
\newblock 2012.
\newblock Models for paired comparison data: A review with emphasis on
  dependent data.
\newblock {\em Statistical Science}  412--433.

\bibitem[\protect\citeauthoryear{Chen and Joachims}{2016}]{chen2016modeling}
Chen, S., and Joachims, T.
\newblock 2016.
\newblock Modeling intransitivity in matchup and comparison data.
\newblock In {\em Proceedings of the Ninth ACM International Conference on Web
  Search and Data Mining},  227--236.
\newblock ACM.

\bibitem[\protect\citeauthoryear{Chiang, Hsieh, and
  Dhillon}{2015}]{chiang2015matrix}
Chiang, K.-Y.; Hsieh, C.-J.; and Dhillon, I.~S.
\newblock 2015.
\newblock Matrix completion with noisy side information.
\newblock In {\em Advances in Neural Information Processing Systems},
  3447--3455.

\bibitem[\protect\citeauthoryear{Copeland}{1951}]{copeland1951reasonable}
Copeland, A.~H.
\newblock 1951.
\newblock A reasonable social welfare function.
\newblock In {\em Mimeographed notes from a Seminar on Applications of
  Mathematics to the Social Sciences, University of Michigan}.

\bibitem[\protect\citeauthoryear{Coppersmith, Fleischer, and
  Rudra}{2006}]{coppersmith2006ordering}
Coppersmith, D.; Fleischer, L.; and Rudra, A.
\newblock 2006.
\newblock Ordering by weighted number of wins gives a good ranking for weighted
  tournaments.
\newblock In {\em Proceedings of the seventeenth annual ACM-SIAM symposium on
  Discrete algorithm},  776--782.
\newblock Society for Industrial and Applied Mathematics.

\bibitem[\protect\citeauthoryear{Ding, Ishwar, and
  Saligrama}{2015}]{ding2015topic}
Ding, W.; Ishwar, P.; and Saligrama, V.
\newblock 2015.
\newblock A topic modeling approach to ranking.
\newblock In {\em AISTATS}.

\bibitem[\protect\citeauthoryear{Jain and Dhillon}{2013}]{jain2013provable}
Jain, P., and Dhillon, I.~S.
\newblock 2013.
\newblock Provable inductive matrix completion.
\newblock {\em arXiv preprint arXiv:1306.0626}.

\bibitem[\protect\citeauthoryear{Ji and Ye}{2009}]{ji2009accelerated}
Ji, S., and Ye, J.
\newblock 2009.
\newblock An accelerated gradient method for trace norm minimization.
\newblock In {\em Proceedings of the 26th annual international conference on
  machine learning},  457--464.
\newblock ACM.

\bibitem[\protect\citeauthoryear{Joachims}{2002}]{joachims2002optimizing}
Joachims, T.
\newblock 2002.
\newblock Optimizing search engines using clickthrough data.
\newblock In {\em Proceedings of the eighth ACM SIGKDD international conference
  on Knowledge discovery and data mining},  133--142.
\newblock ACM.

\bibitem[\protect\citeauthoryear{Kamishima and
  Akaho}{2009}]{kamishima2009efficient}
Kamishima, T., and Akaho, S.
\newblock 2009.
\newblock Efficient clustering for orders.
\newblock In {\em Mining complex data}. Springer.
\newblock  261--279.

\bibitem[\protect\citeauthoryear{Lu and Boutilier}{2011}]{lu2011learning}
Lu, T., and Boutilier, C.
\newblock 2011.
\newblock Learning mallows models with pairwise preferences.
\newblock In {\em Proceedings of the 28th international conference on machine
  learning (icml-11)},  145--152.

\bibitem[\protect\citeauthoryear{Luce}{1959}]{r1959individual}
Luce, R.~D.
\newblock 1959.
\newblock {\em Individual Choice Behavior a Theoretical Analysis}.
\newblock John Wiley and sons.

\bibitem[\protect\citeauthoryear{Marschak and
  others}{1959}]{marschak1959binary}
Marschak, J., et~al.
\newblock 1959.
\newblock Binary choice constraints on random utility indicators.
\newblock Technical report, Cowles Foundation for Research in Economics, Yale
  University.

\bibitem[\protect\citeauthoryear{Rajkumar and Agarwal}{2016}]{rajkumar2016can}
Rajkumar, A., and Agarwal, S.
\newblock 2016.
\newblock When can we rank well from comparisons of $ o (n \log (n)) $
  non-actively chosen pairs?
\newblock In {\em 29th Annual Conference on Learning Theory},  1376--1401.

\bibitem[\protect\citeauthoryear{Thurstone}{1927}]{thurstone1927law}
Thurstone, L.~L.
\newblock 1927.
\newblock A law of comparative judgment.
\newblock {\em Psychological review} 34(4):273.

\bibitem[\protect\citeauthoryear{Xu, Jin, and Zhou}{2013}]{xu2013speedup}
Xu, M.; Jin, R.; and Zhou, Z.-H.
\newblock 2013.
\newblock Speedup matrix completion with side information: Application to
  multi-label learning.
\newblock In {\em Advances in Neural Information Processing Systems},
  2301--2309.

\end{thebibliography}

\section{Proof of Proposition 1}
\begin{proof}
We prove this by showing that every $\mathbf{P} \in \P_n(\psi,r)$ is in $\P_n(\psi,r,\mathbf{A})$ but not the other way around. By the definition of a preference matrix corresponding to the LR model, if $\mathbf{P} \in \P_n(\psi,r)$, then $\text{rank}(\psi(\mathbf{P})) \leq r$. Similarly, for the FLR model, if $\mathbf{P} \in \P_n(\psi,r, \mathbf{A})$, then $\psi(\mathbf{P}) = \mathbf{A}^\top \mathbf{L} \mathbf{A}$ and $\text{rank}(\psi(\mathbf{P})) \leq r$; in other words, $\text{rank}(\mathbf{L}) \leq r$. Now setting $\mathbf{A} = \mathbf{I}$, we have $\P_n(\psi,r) = \P_n (\psi, r, \mathbf{A})$. On the other hand, if $\mathbf{A} \neq \mathbf{I}$, we have $\P_n(\psi,r) \subsetneq \P_n (\psi, r, \mathbf{A})$.
\end{proof}

\section{Proof of Proposition 2}
\begin{proof}
Let $\mathbf{w}$ be the unary score vector in RU models. The result then follows by setting the energy function of item $i$ with respect to item $j$ in the FLR model to be the unary score corresponding to item $i$ in the RU model, ie, by simply setting $\mathbf{F} = \mathbf{I}$ and $\mathbf{W} = \mathbf{0}$ which leads to $E_{ij} = w_i$.
\end{proof}

\section{Proof of Proposition 3}
\begin{proof}
This is immediate by setting $\mathbf{W} = \mathbf{0}$. For concreteness, we choose $\psi$ to be the logit link function. Setting $\mathbf{W} = \mathbf{0}$ in Equation 1
, we obtain
\begin{equation*}
P_{ij} = \frac{e^{-\mathbf{w}^\top \mathbf{f}_i}}{e^{-\mathbf{w}^\top \mathbf{f}_i} + e^{-\mathbf{w}^\top \mathbf{f}_j}}
\end{equation*}
Observe that $\psi(P_{ij}) = \mathbf{w}^\top \mathbf{f}_j - \mathbf{w}^\top \mathbf{f}_i$. Writing this in matrix notation, $\psi(\mathbf{P}) = \mathbf{1} \mathbf{w}^\top \mathbf{F} - \mathbf{F}^\top \mathbf{w} \mathbf{1}^\top$. Note that $\psi(\mathbf{P}) \in \R^{n \times n}$ is a rank-$2$ skew-symmetric matrix. Let $\mathbf{L} := (\mathbf{V} \mathbf{\Sigma}^{-1})^\top \psi(\mathbf{P}) (\mathbf{V} \mathbf{\Sigma}^{-1}) = \mathbf{U}^\top ( \mathbf{1} \mathbf{w}^\top - \mathbf{w} \mathbf{1}^\top ) \mathbf{U}$. Now, note that $\mathbf{L} \in \R^{d \times d}$ is also a rank-$2$ skew-symmetric matrix. Thus, $\mathbf{P} \in \P_n (\text{logit},2,\mathbf{\Sigma V}^\top)$ since $\psi(\mathbf{P}) = (\mathbf{\Sigma V}^\top)^\top \mathbf{L} (\mathbf{\Sigma V}^\top)$. In addition, note that the FLR model accounts for feature correlations when $\mathbf{W} \neq 0$.
\end{proof}

\section{Proof of Lemma 1}
\begin{proof}
For any support $\Xi$, define the following event:
\begin{align*}
\mathfrak{G}_\Xi := \paran{ \abs{\pe_{ij} - P_{ij}} < P_{\min}/2 \quad \forall (i,j) \in \Xi }
\end{align*}
By Hoeffding's bound, $\Pr (\mathfrak{G}_\Xi) \geq 1 - \frac{1}{2n^3}$ whenever $K \geq 11 \log(n)/P_{\min}^2$. Let $L$ be the Lipschitz constant of $\psi$ and set $K \geq \frac{m L^2 \log(n)}{\tau^2}$. Using the inequality that $\frobnorm{\mathbf{N}} \leq \sqrt{m} \infnorm{\mathbf{N}}$, we have
\begin{align*}
& \Pr (\frobnorm{\mathbf{N}} \geq \tau) \leq \Pr \paran{ \infnorm{\mathbf{N}} \geq \frac{\tau}{\sqrt{m}} } \\
& = \Pr \paran{ \exists (i,j) \in \Xi : \abs{ \psi(\pe_{ij}) - \psi(P_{ij}) } \geq \frac{\tau}{\sqrt{m}} } \\
& \leq  \sum_{(i,j) \in \Xi} \Pr \paran{ \abs{ \psi(\pe_{ij}) - \psi(P_{ij}) }  \geq \frac{\tau}{\sqrt{m}} \Big| \mathfrak{G}_\Xi } \Pr (\mathfrak{G}_\Xi) \\ 
&+ \Pr (\mathfrak{G}_\Xi^c) \\
& \leq  \sum_{(i,j) \in \Xi} \Pr \paran{ \abs{ \pe_{ij} - P_{ij} }  \geq \frac{\tau}{L \sqrt{m}} \Big| \mathfrak{G}_\Xi } \Pr (\mathfrak{G}_\Xi)  + \frac{1}{2n^3} \\
& \leq  \sum_{(i,j) \in \Xi} \Pr \paran{ \abs{ \pe_{ij} - P_{ij} }  \geq \frac{\tau}{L \sqrt{m}} } + \frac{1}{2n^3} \\
& \leq \frac{1}{2n^3} + \frac{1}{2n^3} = \frac{1}{n^3}
\end{align*}
\end{proof}

\end{document}